\pdfoutput=1

\documentclass[11pt]{article}

\usepackage[]{ACL2023}

\usepackage{times}
\usepackage{latexsym}

\usepackage[T1]{fontenc}

\usepackage[utf8]{inputenc}

\usepackage{microtype}

\usepackage{inconsolata}

\usepackage{amscd,amsfonts,amsbsy,rotating}
\usepackage{amsthm}
\usepackage{amssymb}
\theoremstyle{definition}
\newtheorem{definition}{Definition}[section]

\usepackage{balance}
\usepackage{subcaption}
\usepackage{graphicx}
\usepackage{epsfig,epstopdf}
\usepackage{multirow}
\usepackage{booktabs}
\usepackage{longtable}
\usepackage{color,xcolor}
\usepackage{url}
\usepackage{latexsym,bm}
\usepackage{enumitem,balance,mathtools}
\usepackage{wrapfig}
\usepackage{euscript}
\usepackage{algorithm}
\usepackage{algorithmic}
\usepackage{ifpdf}
\usepackage{diagbox}
\usepackage{caption}
\usepackage{makecell}
\usepackage[leftcaption]{sidecap}
\usepackage{bm}
\usepackage{textcomp}
\usepackage{upgreek}
\usepackage{booktabs,arydshln}

\makeatletter
\def\adl@drawiv#1#2#3{%
        \hskip.5\tabcolsep
        \xleaders#3{#2.5\@tempdimb #1{1}#2.5\@tempdimb}%
                #2\z@ plus1fil minus1fil\relax
        \hskip.5\tabcolsep}
\newcommand{\cdashlinelr}[1]{%
  \noalign{\vskip\aboverulesep
           \global\let\@dashdrawstore\adl@draw
           \global\let\adl@draw\adl@drawiv}
  \cdashline{#1}
  \noalign{\global\let\adl@draw\@dashdrawstore
           \vskip\belowrulesep}}
\DeclareMathOperator*{\argmax}{arg\,max}

\usepackage{hyperref}
\usepackage{cleveref}
\usepackage{comment}
\crefname{equation}{Eq.}{Eq.}
\crefname{section}{Section}{Sections}
\crefname{appendix}{Appendix}{Appendix}
\crefname{subsection}{Section}{Sections}
\crefname{subsubsection}{Section}{Sections}
\crefname{figure}{Figure}{Figures}
\crefname{table}{Table}{Tables}
\crefname{subfigure}{Figure}{Figures}
\crefname{algocf}{Algorithm}{Algorithms}
\crefname{definition}{Definition}{Definitions}

\title{Targeted Data Generation: Finding and Fixing Model Weaknesses}

\author{Zexue He\Thanks{ Work done during the internship at Microsoft.}\\
  UC San Diego \\ 
  La Jolla, CA, USA\\
  \texttt{zehe@eng.ucsd.edu} \\\And
  Marco Tulio Ribeiro \\
  Microsoft\\
Redmond, WA, USA\\
  \texttt{marcotcr@microsoft.com} \\ \And
  Fereshte Khani \\
  Microsoft\\
Redmond, WA, USA\\
  \texttt{fkhani@microsoft.com}
  }

\begin{document}
\maketitle
\begin{abstract}
Even when aggregate accuracy is high, state-of-the-art NLP models often  fail systematically on specific subgroups of data, resulting in unfair outcomes and eroding user trust. 
Additional data collection may not help in addressing these weaknesses, as such challenging subgroups may be unknown to users, and underrepresented in the existing and new data.
We propose Targeted Data Generation (TDG), a framework that automatically identifies challenging subgroups, and generates new data for those subgroups using large language models (LLMs) with a human in the loop.
TDG estimates the expected benefit and potential harm of data augmentation for each subgroup, and selects the ones most likely to improve within-group performance without hurting overall performance.
In our experiments, TDG\footnote{ Codes and collected data will be released in \url{https://github.com/ZexueHe/TDG}.} significantly improves the accuracy on challenging subgroups for state-of-the-art sentiment analysis and natural language inference models, while also improving overall test accuracy.
\end{abstract}

\section{Introduction}
Despite very high accuracy, state-of-the-art NLP models still exhibit systematic failures on specific subgroups of data.
For example, \citet{rajani2022seal} found that a 95\%-accurate sentiment analysis model did much worse on club reviews (90\%) and movie theater reviews (85\%), while \citet{stuart2018microsoft} notes how a commercial chatbot avoids \emph{any} engagement on topics that even mention Islam or the middle east.
The existence of these \emph{challenging subgroups} can lead to unfair outcomes, erode user trust, and ultimately limit deployment of models, even when \emph{aggregate} accuracy is very high.

One possible solution is to collect or generate more data. However, the additional data may still under-sample from specific challenging subgroups, even if data collection is adversarial \cite{kiela-etal-2021-dynabench}, especially when subgroups are not immediately obvious or salient to humans. Therefore it helps little in addressing these weaknesses.
Tools for discovering challenging subgroups still require human creativity and effort \cite{rajani2022seal}.
\citet{khani2023codev,ribeiro-lundberg-2022-adaptive} show that experts are able to improve existing subgroups via careful data augmentation with large language models (LLMs), but \emph{finding} such challenging subgroups still requires human ingenuity. Perhaps more importantly, they find that naively augmenting certain subgroups can drastically \emph{hurt} other subgroups and overall performance \cite{ribeiro-lundberg-2022-adaptive}.
Hence, the challenge is not only to find challenging subgroups, but also to determine which subgroups are amenable to data augmentation, and how to augment them effectively.

\begin{figure*}[t]
    \centering
\includegraphics[width=\linewidth]{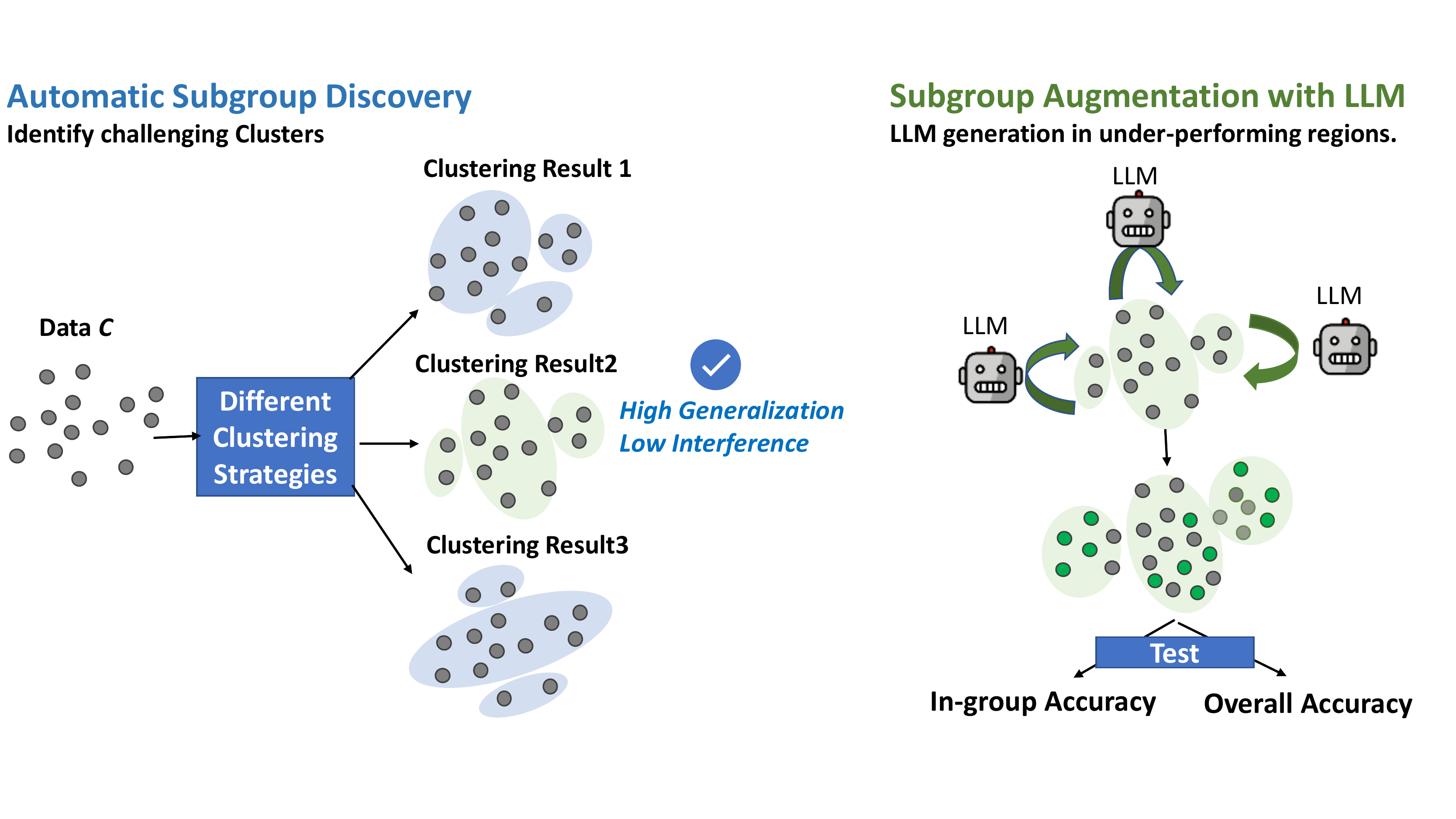}
         \caption{Illustration of the Targeted Data Generation (TDG) pipeline. In the automatic subgroup discovery stage, TDG identifies challenging clusters that can benefit from additional data while minimizing potential negative impacts on performance in other regions (i.e., high generalization (GC) and low interference (IC), as defined in \cref{sec:definition-of--Gc}). In the subgroup augmentation with LLM stage, TDG utilizes GPT-3 to generate additional examples for identified challenging clusters.}
     \label{fig: ours}
\end{figure*}

In this work, we propose Targeted Data Generation (TDG), a framework to automatically identify challenging subgroups that can benefit from more data, and then generate that data with LLMs (\cref{fig: ours}).
Given a target model, TDG clusters validation data into potential challenging subgroups. We then use held-out data to estimate how much each subgroup would benefit from more data, and how much additional data would hurt performance in other regions.
Finally, having identified challenging subgroups amenable to data augmentation, we use GPT-3 \cite{brown2020language} coupled with local subgroup models to generate new data, so as to improve subgroup performance while remaining faithful to the original data distribution.



We evaluate TDG on three tasks: sentiment analysis (SST), paraphrase detection (QQP), and natural language inference (MNLI).
We evaluate various clustering techniques, and find that clustering based on the target model's own representation yields the clusters most amenable to data augmentation (with the exception of QQP, where our analysis indicates label noise would make data augmentation ineffective).
Finally, augmenting these clusters with GPT-3 results in significant improvements on correspondent test clusters, and also small improvements on overall accuracy.
\section{Targeted Data Generation}
Let $\mathcal{M}$ be a target model trained on a training dataset $D_{\text{train}}$, and let $D_{\text{test}}$ be a held-out test dataset. We assume access to a validation dataset $D_{\text{val}}$, which we use to identify and evaluate challenging subgroups. 
We cluster $D_{\text{val}}$ into $k$ disjoint clusters, $\mathcal{C} = \{c_1, c_2, \ldots, c_k\}$, using some clustering technique (we explore various options in  \cref{sec:automatic-discovery-method}, and drop the subscript when talking about a single cluster, for clarity).
We divide $D_{\text{val}}$ randomly into two halves, so that each cluster is divided into $c_{\text{train}}$ and $c_{\text{test}}$ ( $c_{\text{val}}$ can be further divided from $c_{\text{train}}$ if  necessary), to simulate the effect of data augmentation and its impact on the same subgroup. 
We say a cluster $c$ is a \emph{challenging cluster} if the target model $\mathcal{M}$ performs much worse on it than on the overall validation dataset, i.e., $\text{Acc}(\mathcal{M}, c_{\text{train}} \cup c_{\text{val}}) << \text{Acc}(\mathcal{M}, D_{\text{val}})$.

Given a challenging cluster $c$, our goal is to identify whether it is amenable to data augmentation, i.e., more data would generalize and improve performance on $c_{\text{test}}$, without hurting performance on $D_{\text{test}}$.

\subsection{Generalization and Interference, in Context}
\label{sec:definition-of--Gc}
Given the context of $(D_{\text{train}}, \mathcal{M})$ and a target cluster $c$, we obtain a new model $\mathcal{M}'$ by training on a mixture of $D_{\text{train}}$ and $c_{\text{train}}$ (following \citet{ribeiro-lundberg-2022-adaptive}), which effectively upweights examples from $c$ as a surrogate for data augmentation. We use two statistics to evaluate whether $c$ is amenable to data augmentation: Generalization in Context (GC) and Interference in Context (IC).

\begin{definition}[\textbf{Generalization in Context}]  We say a cluster $c$ generalizes in the context of the current model $\mathcal{M}$ and dataset $ D$ if more training on it leads to better performance on hidden examples from the same cluster. Formally, we define Generalization in Context (GC) as

\[ 	\text{GC}(c) = \text{Acc}(\mathcal{M}', c_{\text{val}}) - \text{Acc}(\mathcal{M}, c_{\text{val}}) \]
\end{definition}

GC measures how much the target model can learn from more data from the cluster, and whether that learning transfers to unseen data from the same cluster. A high GC indicates that the cluster is challenging but not hopeless, and that data augmentation could help improve performance. A low GC indicates that the cluster is either already saturated by existing data or too hard for the model to learn, such that more data from the cluster does not help. For example, if the clustering is random, we would expect a low GC, as training on a random subset of data would not improve performance on another random subset. Conversely, if the clustering is based on some meaningful feature that the model struggles with, (such as club reviews \cite{rajani2022seal}), we would expect a high GC, as training on more data from the cluster would help the model overcome its weakness.

\begin{definition}[\textbf{Interference in Context}] We say a cluster $c$ interferes with the original data if augmenting it leads to worse performance on the original data. We could similarly evaluate interference with other clusters, but for now we restrict ourselves to having the original model and dataset as the context. Formally, we define Interference in Context (IC) as

\[ 		\text{IC}(c) = 	\text{Acc}(\mathcal{M}, D_{\text{val}}) - \text{Acc}(\mathcal{M}', D_{\text{val}}) \]
\end{definition}

A high IC indicates that the cluster is incompatible with the original data, and that data augmentation would degrade overall performance. A low IC indicates that the cluster is either similar to the original data, or sufficiently different but not conflicting, such that data augmentation would not hurt overall performance.
For example, if $c$ is label-imbalanced and $D$ is label-balanced, we would expect a high IC, as training on more data from $c$ might bias the model towards a certain label and hurt performance on $D$.
Conversely, if $c$ and $D$ are from different domains but share some common concepts, we would expect a low IC, as training on more data from $c$ would not confuse the model on $D$.
A negative IC indicates that augmenting $c$ actually improves performance on $D$, which could happen if $D$ is small and the model has not saturated it yet, or if there is some domain shift between $D_{\text{test}}$ and $D_{\text{train}}$ which augmentation helps to bridge. 

\paragraph{Aggregate statistics}
To summarize, GC measures whether a cluster benefits from more data, while IC measures whether augmenting that cluster would hurt performance on the original dataset. We aggregate GC and IC over all clusters by taking the average: 

\begin{equation}
    \overline{\text{GC}}(C)=\sum_{i=1}^{k}\frac{\text{GC}(c_i)}{k}\label{eq:def of avg GC}
\end{equation}
\begin{equation}
    \overline {\text{IC}}(C)=\sum_{i=1}^{k}\frac{\text{IC}(c_i)}{k} \label{eq:def of avg }
\end{equation}

\begin{figure*}[h]
    \centering
    \begin{subfigure}{0.25\linewidth}
        \includegraphics[width=0.9\linewidth]{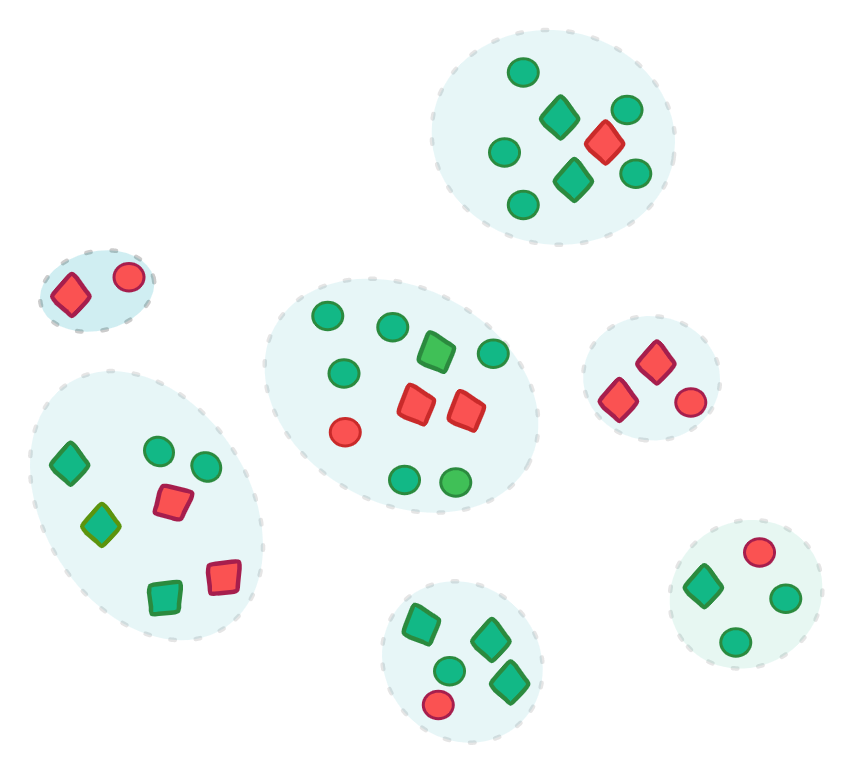}   
        \caption{}
    \end{subfigure}
    \hfill
    \begin{subfigure}{0.32\linewidth}
        \includegraphics[width=\linewidth]{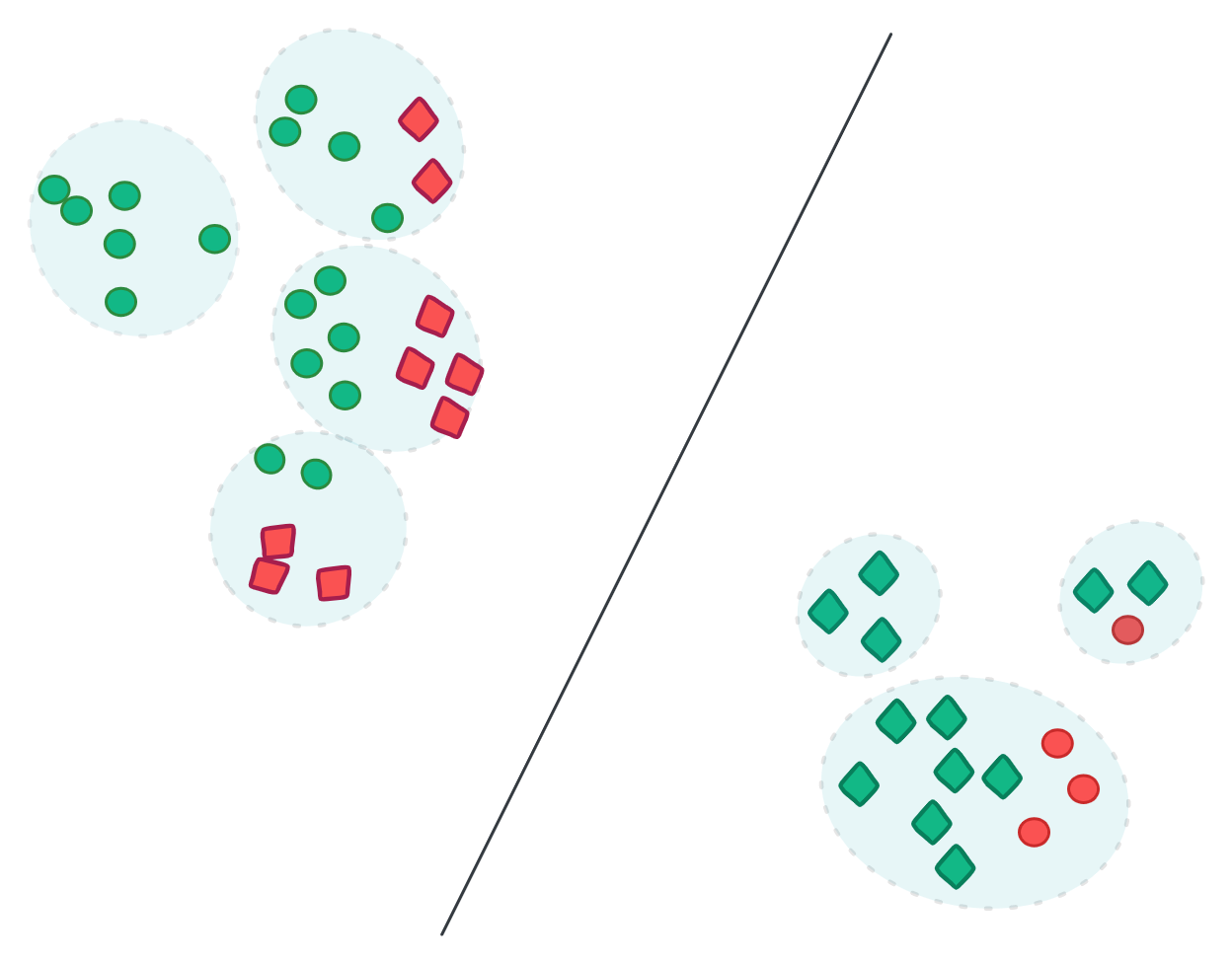}
        \caption{}
    \end{subfigure}
    \hfill
    \begin{subfigure}{0.32\linewidth}
        \includegraphics[width=\linewidth]{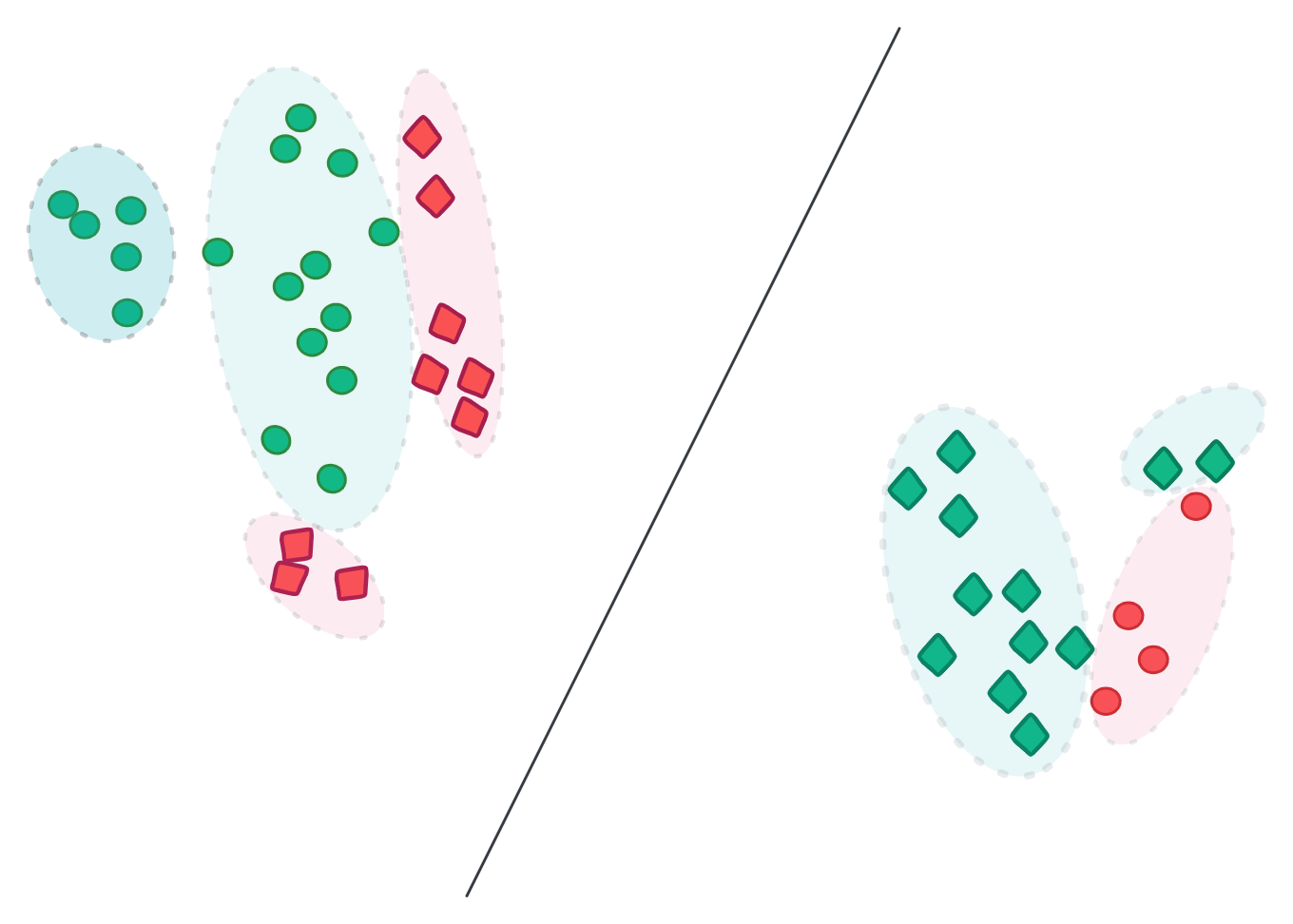}
        \caption{}
    \end{subfigure}
    \hfill
    \caption{Example illustration of cluster results on binary classification from different clustering methods. Data points from binary categories are identified by dots and squares. Errors are shown in red. (a) Agnostic clustering where positive and negative data points are mixed together; (b) Task-based clustering where most points of one category are located at one side of the decision boundary of model $\mathcal{M}$ (being separable by $\mathcal{M}$) and positive/negative points are mixed in clusters; (c) Task-based clustering + label information: besides being separable, data points with the same label can be clustered together.  
    }
    \label{fig:clustering}
\end{figure*}

\subsection{Automatic Subgroup Discovery}
\label{sec:automatic-discovery-method}
We use different representation spaces for clustering, using increasing amounts of information about the task, the model, and the labels. The example is shown in Figure \ref{fig:clustering}. 

\paragraph{Agnostic clustering}
We do not use any information about the task, the model, or the labels, and instead use general-purpose embeddings, such as the embeddings extracted from Sentence-BERT implemented in sentence-transformers \cite{reimers-2019-sentence-bert}, to cluster the validation data. This kind of representations might capture some patterns that the target model cannot currently represent well, and that augmenting these clusters would teach the target model new concepts or relations.

\paragraph{Task-based clustering}
We use the target model's own representation from the second-to-last layer to cluster the validation data. This kind of representations reflects how the target model perceives the data, and might group together examples that the model considers similar or difficult. We expect that if the model relies on spurious correlations or heuristics, these might show up in the representation and get clustered together. Augmenting these clusters would force the model to learn more robust features or strategies.

\paragraph{Task-based + label information}
We use the same representation as task-based clustering, but with the constraint that all examples in a cluster must have the same label (similar to \citet{sohoni2020no}).
While this creates clusters that are clearly label-imbalanced, we expect that examples close in the target representation will also tend to have the same label, and thus this clustering technique should yield clusters with very low or very high error rate (the latter are good candidates as challenging clusters).

\paragraph{Selecting clusters for augmentation}
Given a budget of $k$ clusters we can augment, we evaluate the clustering representations using the aggregate GC and IC statistics of their top-$k$ clusters ranked by error rate, resulting a set of clusters $C_k$.
In other words, we choose a representation that yields the most augmentable clusters without hurting overall performance, as formalized in Equation \ref{topkclusters}.

\begin{equation}
    \mathcal{C}^{*}_{k} = \argmax_{C_k} [\overline {\text{GC}}(C_k) - \overline {\text{IC}}(C_k)] 
\label{topkclusters}
\end{equation}

\subsection{Subgroup Augmentation with LLMs}
In order to augment those top challenging clusters $\mathcal{C}^*_{k}$, we follow the work of \citet{khani2023codev} to use GPT-3 to create similar in-cluster examples, with a human in the loop to provide labels.
We finetune a small local model on each cluster's data and use the disagreement between that model and the current version of $\mathcal{M}'$ to rank GPT-3 generated examples, stopping the process once the current version of the cluster's model mostly agrees with the current version of $\mathcal{M}'$. 
Intuitively, when $\mathcal{M}'$ and the cluster's model converge on cluster data, $\mathcal{M}'$ has learned to generalize to the data in this cluster (thus fulfilling the requirment of GC), and the original $\mathcal{D}$ used when updating $\mathcal{M}'$ should prevent high interference.

\section{Experiments}


\paragraph{Setup} We evaluate the effectiveness of TDG on three tasks from the GLUE benchmark: The Stanford Sentiment Treebank (SST), MultiNLI Matched (MNLI-m)
and Quora Question Pairs (QQP). 
We train a bert-base model for SST and RoBERTa-large models for MNLI and QQP on the official training corpora released in GLUE benchmark to match the best Transformer performance.\footnote{Following \citet{bowman-etal-2015-large, yanaka-etal-2019-neural}, we use the binarized version of MNLI} They are regarded as the target model $\mathcal{M}$ in each task.
We randomly divide the validation data into two half sets: a \textit{dev} set, used for automatic subgroup discovery, and a \textit{devtest} set, used exclusively for evaluation. Therefore, SST has dev size of 436, MNLI dev has size of 4,908, and QQP has dev size of 20,215. We run each experiment five times with different random seeds and report the average scores.

\subsection{Automatic Subgroup Discovery}

\begin{figure*}[t]
\centering
\begin{subfigure}{0.3\textwidth}
    \includegraphics[width=\linewidth]{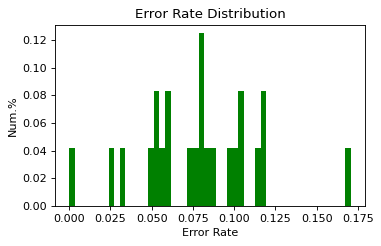}
    \caption{Agnostic clustering; GC=0.0064; IC=0.0000}
         \label{fig:agnostic-sst}
\end{subfigure}
\hfill
\begin{subfigure}{0.3\textwidth}
    \includegraphics[width=\linewidth]{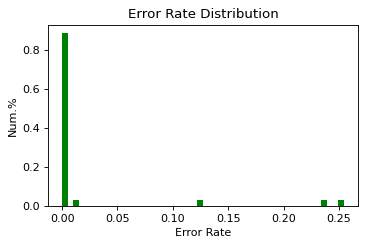}
    \caption{ Task-based clustering; GC=0.011; IC=-0.0002 }
         \label{fig:task-sst}
\end{subfigure}
\hfill
\begin{subfigure}{0.3\textwidth}
    \includegraphics[width=\linewidth]{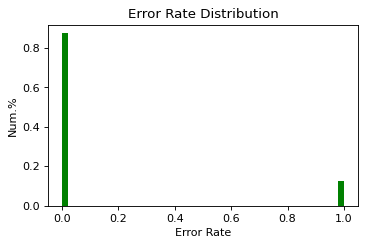}
    \caption{Task-based  + label information; GC=0.1319, IC=0.19298 }
         \label{fig:task-label-sst}
\end{subfigure}
\caption{Error distribution of clusters obtained from three clustering methods on SST. Cluster number k=35. For random clustering: GC=-0.0010, IC=0.0000}
\label{fig:sst_error_analysis}
\end{figure*}
\begin{figure*}[t]
\centering
\begin{subfigure}{0.3\textwidth}
    \includegraphics[width=\linewidth]{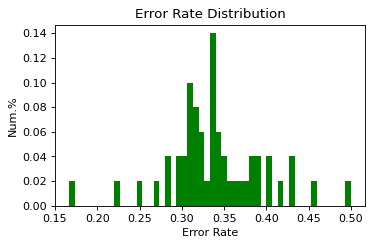}
    \caption{Agnostic clustering; GC=0.0013; IC=0.0011}
         \label{fig:agnostic-mnli}
\end{subfigure}
\hfill
\begin{subfigure}{0.3\textwidth}
    \includegraphics[width=\linewidth]{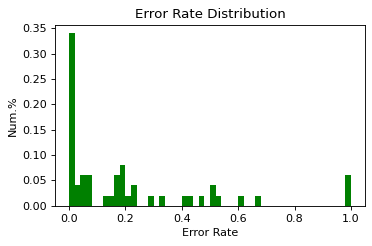}
    \caption{Task-based clustering; GC=0.028; IC=-0.0017 }
         \label{fig:task-mnli}
\end{subfigure}
\hfill
\begin{subfigure}{0.3\textwidth}
    \includegraphics[width=\linewidth]{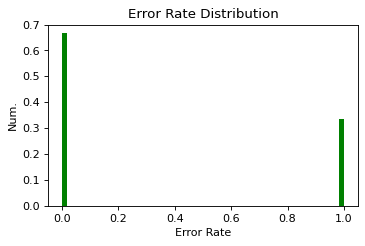}
    \caption{Task-based  + label information; GC=0.0434, IC=0.0023}
         \label{fig:task-label-mnli}
\end{subfigure}
\caption{Error distribution of clusters obtained from three clustering methods on MNLI. Cluster number k=100. For random clustering: GC=-0.0007, IC=0.0002}
\label{fig:mnli_error_analysis}
\end{figure*}

\label{sec: automatic discovery results}
We conduct clustering methods on the dev set of each task. We assign the closest cluster to each instance in the devtest set, such that each cluster in dev has an aligned counterpart for evaluation. We run each clustering method five times using different random seeds and select the clustering results with the best Silhouette scores \cite{rousseeuw1987silhouettes}.  

\paragraph{Comparison of clustering representations}
We present the error rates of discovered clusters  for SST and MNLI in Figures \ref{fig:sst_error_analysis} and \ref{fig:mnli_error_analysis}.
For both tasks, errors were randomly distributed accross clusters produced by agnostic clustering, which indicates that the clusters are not aligned with model behaviors and weaknesses, as also confirmed by the low GC and IC scores. 
In contrast, task-based clustering (with or without label information) results in a large contingent of clusters with zero or few errors (i.e. most successes are clustered together), and a few clusters with higher error rates.
Using label information yields clusters of either all errors or all successes, which results in high Generalization in Context scores, but also high Interference in Context scores. Both are likely due to label imbalance, as we would expect such scores from simply shifting the likelihood of predicting the cluster label.
This analysis thus indicates that task-based clustering without labels yields the clusters that are most amenable to augmentation, since clusters have positive generalization and near-zero interference scores. We use these clusters in subsequent results.

\paragraph{QQP}
All clusterings on QQP (not shown) had very high interference scores, and thus were not deemed suitable for augmentation by TDG.
Indeed, when we piloted data augmentation procedures on these clusters, we saw no tangible benefits.
Manual inspection of clusters indicates that QQP has high label noise (which would explain interference), such that pairs with the same phenomena are often labeled differently, e.g. the pair (``What makes life worth living?'', ``Is life worth it?'') is labeled as  not-duplicate, while (``Why is Deadpool so overrated'', ``Is Deadpool overrated'') is labeled as duplicate.
In this case, TDG correctly identifies a case where subgroup data augmentation is unlikely to be effective, and other solutions (e.g. data cleaning) should be pursued.
We do not report any QQP results from now on.

\subsection{Subgroup Augmentation with LLMs}
\label{sec: TDG performance}
Based on the high-GC and low-IC clusters discovered in previous step, we conduct augmentation targeted on those clusters with large language models with human in the loop.

\paragraph{Human Participants} We recruited 12 users to label GPT-3 generated data in the subgroup augmentation step. All users are from academia or industry (with IRB approval) and have experience working with AI-based natural language generation systems (e.g. GPT-3).
Each user was assigned a high-error cluster discovered in the automatic subgroup discovery step (2 from SST and 10 from MNLI), and asked to label GPT-3 generations. We use the original sentences from the cluster as the initial prompt.
Sentences that users labeled differently from the model's prediction were added to the augmented set. We allocated 90 minutes for user labeling, with more information in the Appendix \ref{sec:appendix-hiip}.

\paragraph{Baselines}We compare TDG to the following previous works that aim to improve subgroup performance: (1) \textbf{Reweighing} \cite{sohoni2020no}, which addresses hidden stratification caused by dataset imbalance by optimizing the per-cluster worst-case performance. In our experiments, we use the same Group Distributionally Robust Optimization (GDRO) introduced in their work on each cluster as the fine-tuning objective. (2) \textbf{Paraphrasing} where we use Parrot \cite{prithivida2021parrot}, a T5-based paraphrase model, to generate similar examples of data points in clusters as an augmentation. The size of the final fine-tune set is the same as TDG for a fair comparison.



\paragraph{One cluster at a time v.s. simultaneous augmentation}
Each participant augmented a single cluster, and we report these results as \textbf{TDG(single)}, noting that for these we only measure in-cluster performance.  We further  pool the data from all participants (\textbf{TDG(all)}) to test the improvements on each cluster as well as performance on the overall test set (devtest).
In each experiment, in order to avoid the issue of catastrophic forgetting \cite{mccloskey1989catastrophic}, we randomly sampled training data with the same frequency as TDG augmented data in the fine-tuning process\footnote{In MNLI experiment, due to the high interference among clusters,  we adjust the weights of  training samples and collected responses when combining all data points for TDG(all) in fine-tuning (i.e.,  we set portions of original samples:user responses = 2:1). In SST, all responses are combined without any adjustment.}.

\begin{table}[h]
\centering
\resizebox{\columnwidth}{!}{%
\begin{tabular}{@{}ccccc@{}}
\toprule
\multirow{2}{*}{\textbf{Model}} & \multicolumn{4}{c}{\textbf{SST}}                                                  \\ \cmidrule(l){2-5} 
                                & \textbf{1st}  & \textbf{2nd}   & \textbf{Avg Cluster} & \textbf{devtest} \\ \midrule
\textbf{BERT-base}          & 81.74         & 81.13          & 81.45                & 93.77            \\ \cmidrule(lr){1-5}
\textbf{Reweighing}             & 78.7          & 82.03          & 80.37                & 93.49            \\
\textbf{Paraphrasing}           & 77.61         & 82.42          & 80.02                & 92.26            \\\cmidrule(lr){1-5}
\textbf{TDG (single)}                    & \textbf{83.8} & \textbf{83.39} & \textbf{83.60}      & -                \\
\textbf{TDG (all)}              & 82.61         & \textbf{83.39} & 83.00               & \textbf{94.32}   \\ \bottomrule
\end{tabular}%
}
\caption{Accuracy of TDG v.s. baselines tested on top-2 error clusters and left-out devtest set of SST. BERT-base is the target model $\mathcal{M}$.}
\label{tab:accuracy-baseline-sst}
\end{table}

\begin{table*}[ht]
\centering
\resizebox{\textwidth}{!}{%
\begin{tabular}{@{}ccccccccccccc@{}}
\toprule
\multirow{2}{*}{\textbf{Model}} & \multicolumn{12}{c}{\textbf{MNLI}}                                                             \\ \cmidrule(l){2-13} 
 &
  \textbf{1st} &
  \textbf{2nd} &
  \textbf{3rd} &
  \textbf{4th} &
  \textbf{5th} &
  \textbf{6th} &
  \textbf{7th} &
  \textbf{8th} &
  \textbf{9th} &
  \textbf{10th} &
  \textbf{Avg Cluster} &
  \textbf{devtest} \\ \midrule
\textbf{RoBERTa-Large}          & 51.85 & \textbf{53.57} & 53.85 & 54.84 & 55.56 & 58.82 & 65.71 & 66.56 & \textbf{68.75} & 76.19 & 60.57  & 93.46 \\
\textbf{Reweighing}             & 51.85 & \textbf{53.57} & 30.77 & 58.06 & 55.56 & 58.82 & 68.57 & 65.91 & \textbf{68.75} & 73.81 & 58.57 & 93.46 \\
\textbf{Paraphrasing}           & 51.85 & 42.86 & 53.85 & 54.84 & 44.44 & 58.82 & 65.71 & 65.91 & \textbf{68.75} & 26.19 & 53.32 & 86.45 \\
\textbf{TDG (single)} &
  51.85 &
  \textbf{53.57} &
  61.54 &
  \textbf{67.74} &
  \textbf{66.67} &
  \textbf{64.71} &
  65.71 &
  \textbf{75.68} &
  66.67 &
  76.19 &
  \textbf{65.03} &
  - \\
\textbf{TDG (all)} &
  \textbf{59.26} &
  \textbf{53.57} &
  \textbf{64.28} &
  61.29 &
  55.56 &
  \textbf{64.71} &
  \textbf{74.28} &
  68.18 &
  \textbf{68.75} &
  \textbf{78.57} &
  64.85 &
  \textbf{93.62} \\ \bottomrule
\end{tabular}%
}
\caption{Accuracy of different models tested on top-10 high-error clusters and left-out devtest set of  MNLI.}
\label{tab:accuracy-baseline-mnli}
\end{table*}


\paragraph{Improvement in challenging subgroups}  
\cref{tab:accuracy-baseline-sst} and \cref{tab:accuracy-baseline-mnli} show the results of all baselines, as well as TDG(single) and the aggregated TDG(all), on the SST and MNLI tasks, respectively. For both tasks, augmenting individual clusters with TDG tends to be more effective than all baselines and ablations, as the average in-cluster accuracy has been increased from 81.45\% to 83.60\% on SST and from 60.57\% to 65.03\% on MNLI, which is higher than any baseline models. Additionally, we also observed that adding TDG data from all clusters can improve all clusters by an average of 4.28\% (from 60.57\% to 64.85\%) on MNLI and an average of 1.55\% (from 81.45\% to 83.00\%) on SST, which is also higher than all baseline models.  
Note that the accuracy of every single cluster in TDG(all) is better than the target model. 
For some challenging clusters, augmentation on their own (TDG(single)) may yield better results, due to potential interference between clusters (see Appendix \ref{sec:cluster-relationship} for more details). 

\paragraph{Improvement in overall devtest} 
We observed an improvement in overall performance on the devtest set with TDG(all), with an increase of 0.55\% on SST and 0.16\% on MNLI. This suggests that improving challenging clusters has the potential to improve the model at a global level, while neither baselines were able to achieve this. We notice the improvement on the devtest set is not as significant as the improvement on individual low-performed groups. This is likely due to the fact that these vulnerable groups are usually minorities and their representation in the devtest set is small (e.g., the average size of the 10 clusters in MNLI experiment is just 88 whereas the devtest has size of 4,908), diluting the impact of the improvement. 

 \paragraph{Ablation Analysis}
 We  evaluate the following variations of TDG to test the effectiveness of each step: 
 \begin{itemize}
 \item \textbf{Automatic  Subgroup Discovery Only} in which the fine-tuning data is created by using the same clusters as TDG but without augmentation  and adding the same number of random samples from the training data, to test the error discovery step. 
 \item \textbf{Subgroup Augmentation with LLM Only} in which the fine-tuning data is created by using $n$ random samples from the dev set ($n$ is the number of total sentences in challenging clusters used in TDG) and applying subgroup augmentation with GPT-3, to test the effectiveness of the augmentation. Augmentation ends once the same number of augmented data as TDG is reached.
 \end{itemize}
\begin{table}[h]
\centering
\resizebox{\linewidth}{!}{%
\begin{tabular}{ccccccc}
\toprule
\multirow{2}{*}{\textbf{Model}} & \multicolumn{4}{c}{\textbf{SST}}                                          \\ \cmidrule(l){2-5} 
                                & \textbf{1st}   & \textbf{2nd}   & \textbf{Avg Cluster} & \textbf{devtest} \\ \midrule
\textbf{BERT-base}              & 81.74          & 81.13          & 81.45                & 93.77            \\\cmidrule(lr){1-5}
\textbf{\begin{tabular}[c]{@{}c@{}}Automatic Subgroup \\ Discovery only\end{tabular}}   & 78.70 & 82.20 & 80.45 & 93.89 \\
\textbf{\begin{tabular}[c]{@{}c@{}}Subgroup Augmentation \\ with LLM only\end{tabular}} & 79.42 & 78.42 & 78.91 & 93.17 \\\cmidrule(lr){1-5}
\textbf{TDG (single)}           & \textbf{83.80} & \textbf{83.39} & \textbf{83.60}      & -                \\
\textbf{TDG (all)}              & 82.61          & \textbf{83.39} & 83.00               & \textbf{94.32}   \\ \bottomrule
\end{tabular}%
}
\caption{
Accuracy of different ablations of TDG on top-2 high-error clusters in SST. BERT-base is the target model $\mathcal{M}$.}
\label{tab:ablation-sst}
\end{table}

We see that fine-tuning with clusters alone can improve performance on certain clusters when the size is sufficient (e.g., 2nd in SST), but it can also lead to over-fitting and reduced performance (e.g., 1st in SST). 
Additionally, subgroup augmentation on randomly sampled clusters results in a decrease in performance not only in low-performing areas, but also overall on the devtest set. Without the automatic subgroup discovery, the GPT-3 augmented sentences may introduce more noise rather than benefits, which verifies the bottleneck of previous work \cite{ribeiro-lundberg-2022-adaptive} and emphasizes the importance of the automatic subrgoup discovery.
\begin{table*}[t]
\centering
\resizebox{\textwidth}{!}{%
\begin{tabular}{@{}llcc@{}}
\toprule
 &
  \multicolumn{1}{c}{\textbf{\textit{Cluster: Having multiple sentiments and one is dominating than the rest}}} &
  \textbf{Label} &
  \textbf{Prediction} \\  \cmidrule(r){2-4} 
\multirow{2}{*}{\textbf{SST}} &
  \begin{tabular}[c]{@{}l@{}}On the heels of the ring comes a similarly morose and humorless horror movie that, \\ although flawed , is to be commended for its straight-ahead approach to creepiness .\end{tabular} &
  positive &
  negative \\\cdashlinelr{2-4}
 &
 \begin{tabular}[c]{@{}l@{}}Another one of those estrogen overdose movies like "divine secrets of the ya ya sisterhood ''\\ except that the writing , acting and character development are a lot better .\end{tabular} &
  positive &
  negative \\ \midrule
\multirow{20}{*}{\textbf{MNLI}} &
  \multicolumn{1}{c}{\textbf{\textit{Cluster: Having same meaning. Formal Tone v.s. Casual Tone}}} &
  \textbf{Label} &
  \textbf{Prediction} \\ \cmidrule(l){2-4}
 &
  \begin{tabular}[c]{@{}l@{}}\textbf{Sentence1}: Do you think I should be concerned? \\ \textbf{Sentence2}: Do you think it is a problem\end{tabular} &
  entailment & 
  \begin{tabular}[c]{@{}c@{}}not \\ enatilment\end{tabular} \\\cdashlinelr{2-4}
 &
  \begin{tabular}[c]{@{}l@{}}\textbf{Sentence1}: He seemed too self-assured. \\ \textbf{Sentence2}: He is very cocky\end{tabular} &
  entailment &
  \begin{tabular}[c]{@{}c@{}}not \\ enatilment\end{tabular} \\ \cmidrule(r){2-4} 
 &
  \multicolumn{1}{c}{\textit{\textbf{Cluster: One v.s. All}}} &
  \textbf{Label} &
  \textbf{Prediction} \\ \cmidrule(rl){2-4}
 &
  \begin{tabular}[c]{@{}l@{}}\textbf{Sentence1}: Pray be seated, mademoiselle. \\ \textbf{Sentence2}: Please, everyone be seated.\end{tabular} &
   \begin{tabular}[c]{@{}c@{}}not \\ enatilment\end{tabular}  & entailment
  \\\cdashlinelr{2-4}
 &
  \begin{tabular}[c]{@{}l@{}}\textbf{Sentence1}: Similar conclusions have been reached by legal studies in a dozen states including Florida. \\ \textbf{Sentence2}: Similar conclusions have been seen across the world. \end{tabular} & \begin{tabular}[c]{@{}c@{}}not \\ enatilment\end{tabular}
  & entailment
   \\ \cmidrule(r){2-4} 
 &
 \multicolumn{1}{c}{\textbf{\textit{Cluster: Suspicion v.s. Fact}}} &
  \textbf{Label} &
  \textbf{Prediction} \\\cmidrule(l){2-4}
 &
  \begin{tabular}[c]{@{}l@{}}\textbf{Sentence1}: The analysis also addresses the various alternatives to the final rule  which were considered, \\including differing compliance or reporting requirements,  use of performance rather than design standards,\\ and an exemption for small entities from coverage of the rule. \\ \textbf{Sentence2}: The rule is subject to change."\end{tabular} &
  \begin{tabular}[c]{@{}c@{}}not \\ enatilment\end{tabular} &
  entailment \\ \cdashlinelr{2-4}
 &
  \begin{tabular}[c]{@{}l@{}}\textbf{Sentence1}: In the depths of the Cold War, many Americans suspected Communists \\ had infiltrated Washington and were about to subvert our democracy. \\ \textbf{Sentence2}: Communists infiltrated Washington during the Cold War.\end{tabular} &
  \begin{tabular}[c]{@{}c@{}}not \\ enatilment\end{tabular} &
  entailment \\ \bottomrule
\end{tabular}%
}
\caption{Interpretation about discovered high-error clusters. Each cluster is shown with two errors.}
\label{tab:interpretation-on-cluster-sentence}
\end{table*}

\paragraph{Interpretation of low-performed groups}
In this section, we present some examples from the high-error groups discovered in automatic subgroup discovery. We also provide readable interpretations for the clusters as shown in \cref{tab:interpretation-on-cluster-sentence}. Our automatic subgroup discovery is able to identify meaningful errors, such as mis-identifying the dominant sentiment from a mixture of sentiments in SST, or errors related to different language tones in MNLI. Furthermore, we also notice complex patterns in reasoning is identified, such as Factivity and Monotonicity, which are recognized challenges in SuperGLUE Diagnostic tasks.



\section{Related Work}
Recent research in machine learning has focused on enhancing the robust performance of models by identifying challenging subgroups and improving their performance. 

\noindent\textbf{Discovering Challenging Subgroups}
Several studies, such as \citet{d2022spotlight} and \citet{rajani2022seal}, focus on identifying challenging subgroups in the data. However, these works primarily focus on discovering general low-performing regions in embedding space and do not address strategies for improving these regions. In contrast, our work aims to identify challenging subgroups that are also amenable to improvement through data augmentation using language models.

\noindent\textbf{Improving Performance of Known Subgroups}
Other studies, such as \citet{thakur2021augmented,yoo2021gpt3mix,he2021detect}, focus on augmenting data from known subgroups or patterns. However, it can be challenging to apply these methods in scenarios where the challenging subgroups are not known a priori. Another stream of work focuses on model testing and debugging, which involves creating human-generated data points and testing them on the model. Methods such as CheckList \cite{ribeiro2020beyond} and DynaBench \cite{kiela-etal-2021-dynabench} generate test cases from pre-defined topics and templates, while AdaTest \cite{ribeiro-lundberg-2022-adaptive} uses pre-trained language models to generate more tests that are similar to the human-created examples.
Although these methods show promising results in improving the performance of challenging subgroups, it is not clear how to provide the first data points from a challenging subgroup. Finding such data points was the main focus of our work, where we showed how to find data points that are suitable for further augmentation.

\noindent\textbf{Model-based Approaches}
Another approach for enhancing the performance of challenging subgroups is to develop new training strategies. \citet{sagawa2019distributionally} minimize the worst group accuracy when subgroups are known a priori, \citet{khani2019maximum} add variance of loss to the optimization function, and \citet{liu2021just} train the model twice, one with every data point and once more with the ones that have high losses. \citet{sohoni2020no} discovered subgroups and then change the training function to improve the accuracy. Changing the training function usually improves the accuracy of challenging subgroups, but at the expense of decreasing accuracy in other subgroups or the overall accuracy. In contrast, our work increases the performance of challenging subgroups while also increasing the overall accuracy.

\paragraph{Data Augmentation with Human-in-The-Loop} Recent works note that Human-in-The-Loop (HITL) based augmentation offers unique benefits over automatic data augmentation, such as addressing dataset design flaws \cite{fanton-etal-2021-human}, improving performance for minority groups \cite{srivastava2020robustness}, and avoiding syntactic and semantic distortions in the text \cite{anaby2020not}.

We want to point out that TDG is orthogonal to non-HITL augmentation (i.e. they can be used together). In addition, TDG's use of LLM to generate augmentations for specific data groups helps reduce the human effort – TDG only requires minimal human effort for validation, making it more efficient than previous HITL-based methods that either require domain experts or require more extensive human input. In this paper, we purposefully chose state-of-the-art (SOTA) models that are already very good. However, our work shows that even such models still exhibit coherent lower-performance groups that can be further improved with targeted data collection.

\section{Conclusion}
In this work, we presented a thorough analysis of error distribution among different groups and introduced Targeted Data Generation (TDG), a framework that automatically identifies challenging groups that are amenable to improvement through data augmentation using large language models (LLMs) without negatively impacting overall accuracy. Our experiments with state-of-the-art models demonstrate that TDG is able to improve in-group performance by 2-13\% while also increasing overall accuracy. Furthermore, TDG was able to improve performance for every single selected cluster without interference, indicating its potential as a reliable approach for a new data collection framework. As LLMs continue to advance and are trained on more diverse and large corpora, TDG represents a promising approach for addressing the weaknesses of simpler models.

\section{Ethic Considerations}
In this paper, we propose a method for automatically identifying groups of data that are underperforming due to a lack of training examples. It is important to note that these underperforming groups may be related to marginalized demographic groups, which may be underrepresented in the data. By identifying these groups, our work is able to reveal potential discriminatory behaviors in NLP models and facilitate bias mitigation by augmenting these underrepresented groups. However, there is also the risk that malicious actors may exploit this information and create adversarial examples that further bias the model. To address this concern, we suggest involving the user audience or implementing fairness regulations in the interactive procedure to prevent such behaviors.
Finally, it's worth noting that our model relies heavily on large language models to improve the performance of challenging groups as a result if some groups are not represented in LLMs our method is unable to increase their performance.  

\section{Limitations}
One limitation of our approach is that we aggregated IC and GC measurements over clusters during the automatic subgroup discovery process, but we did not fully consider the relationships between clusters. A more comprehensive strategy for utilizing beneficial relationships and a more precise approach to potential conflicts between clusters could lead to further improvements in  overall performance. Additionally, our MNLI experiments were conducted on large dataset that had multiple clusters with errors. We chose to focus on the top-10 clusters with the most errors due to limitations in resources for running a user study. While TDG on top-K clusters has demonstrated effectiveness in improving performance, there is still the potential for further improvements by working on a larger number of clusters. At the same time, we emphasize that TDG should be used as the last step to improve performance in low-performing groups (clusters with high errors). If these groups are numerous, it means the model is likely under-trained, and other techniques (e.g. better data/modeling) should be applied first.

\section{Acknowledgements}
We would like to thank Scott Lundberg for his kind assistance in designing and implementing the user interface. We are appreciative of the insightful suggestions provided by folks from the Microsoft  Office of Applied Research. Special thanks go to Brent Hecht, Aaron Halfaker, and Yujin Kim for their generous contributions of time and support in our user studies. We would also like to express our thanks to all the participants from the University of California San Diego for their active involvement in the user studies.

\bibliography{anthology,custom}

\begin{thebibliography}{24}
\expandafter\ifx\csname natexlab\endcsname\relax\def\natexlab#1{#1}\fi

\bibitem[{Anaby-Tavor et~al.(2020)Anaby-Tavor, Carmeli, Goldbraich, Kantor,
  Kour, Shlomov, Tepper, and Zwerdling}]{anaby2020not}
Ateret Anaby-Tavor, Boaz Carmeli, Esther Goldbraich, Amir Kantor, George Kour,
  Segev Shlomov, Naama Tepper, and Naama Zwerdling. 2020.
\newblock Do not have enough data? deep learning to the rescue!
\newblock In \emph{Proceedings of the AAAI Conference on Artificial
  Intelligence}, volume~34, pages 7383--7390.

\bibitem[{Bowman et~al.(2015)Bowman, Angeli, Potts, and
  Manning}]{bowman-etal-2015-large}
Samuel~R. Bowman, Gabor Angeli, Christopher Potts, and Christopher~D. Manning.
  2015.
\newblock \href {https://doi.org/10.18653/v1/D15-1075} {A large annotated
  corpus for learning natural language inference}.
\newblock In \emph{Proceedings of the 2015 Conference on Empirical Methods in
  Natural Language Processing}, pages 632--642, Lisbon, Portugal. Association
  for Computational Linguistics.

\bibitem[{Brown et~al.(2020)Brown, Mann, Ryder, Subbiah, Kaplan, Dhariwal,
  Neelakantan, Shyam, Sastry, Askell et~al.}]{brown2020language}
Tom~B Brown, Benjamin Mann, Nick Ryder, Melanie Subbiah, Jared Kaplan, Prafulla
  Dhariwal, Arvind Neelakantan, Pranav Shyam, Girish Sastry, Amanda Askell,
  et~al. 2020.
\newblock Language models are few-shot learners. arxiv 2020.
\newblock \emph{arXiv preprint arXiv:2005.14165}, 4.

\bibitem[{Damodaran(2021)}]{prithivida2021parrot}
Prithiviraj Damodaran. 2021.
\newblock Parrot: Paraphrase generation for nlu.

\bibitem[{d'Eon et~al.(2022)d'Eon, d'Eon, Wright, and
  Leyton-Brown}]{d2022spotlight}
Greg d'Eon, Jason d'Eon, James~R Wright, and Kevin Leyton-Brown. 2022.
\newblock The spotlight: A general method for discovering systematic errors in
  deep learning models.
\newblock In \emph{2022 ACM Conference on Fairness, Accountability, and
  Transparency}, pages 1962--1981.

\bibitem[{Fanton et~al.(2021)Fanton, Bonaldi, Tekiro{\u{g}}lu, and
  Guerini}]{fanton-etal-2021-human}
Margherita Fanton, Helena Bonaldi, Serra~Sinem Tekiro{\u{g}}lu, and Marco
  Guerini. 2021.
\newblock \href {https://doi.org/10.18653/v1/2021.acl-long.250}
  {Human-in-the-loop for data collection: a multi-target counter narrative
  dataset to fight online hate speech}.
\newblock In \emph{Proceedings of the 59th Annual Meeting of the Association
  for Computational Linguistics and the 11th International Joint Conference on
  Natural Language Processing (Volume 1: Long Papers)}, pages 3226--3240,
  Online. Association for Computational Linguistics.

\bibitem[{He et~al.(2021)He, Majumder, and McAuley}]{he2021detect}
Zexue He, Bodhisattwa~Prasad Majumder, and Julian McAuley. 2021.
\newblock Detect and perturb: Neutral rewriting of biased and sensitive text
  via gradient-based decoding.
\newblock In \emph{Findings of the Association for Computational Linguistics:
  EMNLP 2021}, pages 4173--4181.

\bibitem[{Khani et~al.(2019)Khani, Raghunathan, and Liang}]{khani2019maximum}
Fereshte Khani, Aditi Raghunathan, and Percy Liang. 2019.
\newblock Maximum weighted loss discrepancy.
\newblock \emph{arXiv preprint arXiv:1906.03518}.

\bibitem[{Khani and Ribeiro(2023)}]{khani2023codev}
Fereshte Khani and Marco~Tulio Ribeiro. 2023.
\newblock Collaborative development of nlp models.
\newblock \emph{arXiv preprint arXiv:2305.12219}.

\bibitem[{Kiela et~al.(2021)Kiela, Bartolo, Nie, Kaushik, Geiger, Wu, Vidgen,
  Prasad, Singh, Ringshia, Ma, Thrush, Riedel, Waseem, Stenetorp, Jia, Bansal,
  Potts, and Williams}]{kiela-etal-2021-dynabench}
Douwe Kiela, Max Bartolo, Yixin Nie, Divyansh Kaushik, Atticus Geiger,
  Zhengxuan Wu, Bertie Vidgen, Grusha Prasad, Amanpreet Singh, Pratik Ringshia,
  Zhiyi Ma, Tristan Thrush, Sebastian Riedel, Zeerak Waseem, Pontus Stenetorp,
  Robin Jia, Mohit Bansal, Christopher Potts, and Adina Williams. 2021.
\newblock \href {https://doi.org/10.18653/v1/2021.naacl-main.324} {Dynabench:
  Rethinking benchmarking in {NLP}}.
\newblock In \emph{Proceedings of the 2021 Conference of the North American
  Chapter of the Association for Computational Linguistics: Human Language
  Technologies}, pages 4110--4124, Online. Association for Computational
  Linguistics.

\bibitem[{Liu et~al.(2021)Liu, Haghgoo, Chen, Raghunathan, Koh, Sagawa, Liang,
  and Finn}]{liu2021just}
Evan~Z Liu, Behzad Haghgoo, Annie~S Chen, Aditi Raghunathan, Pang~Wei Koh,
  Shiori Sagawa, Percy Liang, and Chelsea Finn. 2021.
\newblock Just train twice: Improving group robustness without training group
  information.
\newblock In \emph{International Conference on Machine Learning}, pages
  6781--6792. PMLR.

\bibitem[{McCloskey and Cohen(1989)}]{mccloskey1989catastrophic}
Michael McCloskey and Neal~J Cohen. 1989.
\newblock Catastrophic interference in connectionist networks: The sequential
  learning problem.
\newblock In \emph{Psychology of learning and motivation}, volume~24, pages
  109--165. Elsevier.

\bibitem[{Rajani et~al.(2022)Rajani, Liang, Chen, Mitchell, and
  Zou}]{rajani2022seal}
Nazneen Rajani, Weixin Liang, Lingjiao Chen, Meg Mitchell, and James Zou. 2022.
\newblock Seal: Interactive tool for systematic error analysis and labeling.
\newblock \emph{arXiv preprint arXiv:2210.05839}.

\bibitem[{Reimers and Gurevych(2019)}]{reimers-2019-sentence-bert}
Nils Reimers and Iryna Gurevych. 2019.
\newblock \href {http://arxiv.org/abs/1908.10084} {Sentence-bert: Sentence
  embeddings using siamese bert-networks}.
\newblock In \emph{Proceedings of the 2019 Conference on Empirical Methods in
  Natural Language Processing}. Association for Computational Linguistics.

\bibitem[{Ribeiro and Lundberg(2022)}]{ribeiro-lundberg-2022-adaptive}
Marco~Tulio Ribeiro and Scott Lundberg. 2022.
\newblock \href {https://doi.org/10.18653/v1/2022.acl-long.230} {Adaptive
  testing and debugging of {NLP} models}.
\newblock In \emph{Proceedings of the 60th Annual Meeting of the Association
  for Computational Linguistics (Volume 1: Long Papers)}, pages 3253--3267,
  Dublin, Ireland. Association for Computational Linguistics.

\bibitem[{Ribeiro et~al.(2020)Ribeiro, Wu, Guestrin, and
  Singh}]{ribeiro2020beyond}
Marco~Tulio Ribeiro, Tongshuang Wu, Carlos Guestrin, and Sameer Singh. 2020.
\newblock Beyond accuracy: Behavioral testing of nlp models with checklist.
\newblock In \emph{Proceedings of the 58th Annual Meeting of the Association
  for Computational Linguistics}, pages 4902--4912.

\bibitem[{Rousseeuw(1987)}]{rousseeuw1987silhouettes}
Peter~J Rousseeuw. 1987.
\newblock Silhouettes: a graphical aid to the interpretation and validation of
  cluster analysis.
\newblock \emph{Journal of computational and applied mathematics}, 20:53--65.

\bibitem[{Sagawa et~al.(2019)Sagawa, Koh, Hashimoto, and
  Liang}]{sagawa2019distributionally}
Shiori Sagawa, Pang~Wei Koh, Tatsunori~B Hashimoto, and Percy Liang. 2019.
\newblock Distributionally robust neural networks.
\newblock In \emph{International Conference on Learning Representations}.

\bibitem[{Sohoni et~al.(2020)Sohoni, Dunnmon, Angus, Gu, and
  R{\'e}}]{sohoni2020no}
Nimit Sohoni, Jared Dunnmon, Geoffrey Angus, Albert Gu, and Christopher R{\'e}.
  2020.
\newblock No subclass left behind: Fine-grained robustness in coarse-grained
  classification problems.
\newblock \emph{Advances in Neural Information Processing Systems},
  33:19339--19352.

\bibitem[{Srivastava et~al.(2020)Srivastava, Hashimoto, and
  Liang}]{srivastava2020robustness}
Megha Srivastava, Tatsunori Hashimoto, and Percy Liang. 2020.
\newblock Robustness to spurious correlations via human annotations.
\newblock In \emph{International Conference on Machine Learning}, pages
  9109--9119. PMLR.

\bibitem[{Stuart-Ulin(2018)}]{stuart2018microsoft}
Chloe~Rose Stuart-Ulin. 2018.
\newblock Microsoft’s politically correct chatbot is even worse than its
  racist one.
\newblock \emph{Quartz Ideas}, 31.

\bibitem[{Thakur et~al.(2021)Thakur, Reimers, Daxenberger, and
  Gurevych}]{thakur2021augmented}
Nandan Thakur, Nils Reimers, Johannes Daxenberger, and Iryna Gurevych. 2021.
\newblock Augmented sbert: Data augmentation method for improving bi-encoders
  for pairwise sentence scoring tasks.
\newblock In \emph{Proceedings of the 2021 Conference of the North American
  Chapter of the Association for Computational Linguistics: Human Language
  Technologies}, pages 296--310.

\bibitem[{Yanaka et~al.(2019)Yanaka, Mineshima, Bekki, Inui, Sekine,
  Abzianidze, and Bos}]{yanaka-etal-2019-neural}
Hitomi Yanaka, Koji Mineshima, Daisuke Bekki, Kentaro Inui, Satoshi Sekine,
  Lasha Abzianidze, and Johan Bos. 2019.
\newblock \href {https://doi.org/10.18653/v1/W19-4804} {Can neural networks
  understand monotonicity reasoning?}
\newblock In \emph{Proceedings of the 2019 ACL Workshop BlackboxNLP: Analyzing
  and Interpreting Neural Networks for NLP}, pages 31--40, Florence, Italy.
  Association for Computational Linguistics.

\bibitem[{Yoo et~al.(2021)Yoo, Park, Kang, Lee, and Park}]{yoo2021gpt3mix}
Kang~Min Yoo, Dongju Park, Jaewook Kang, Sang-Woo Lee, and Woomyeong Park.
  2021.
\newblock Gpt3mix: Leveraging large-scale language models for text
  augmentation.
\newblock \emph{arXiv preprint arXiv:2104.08826}.

\end{thebibliography}
\bibliographystyle{acl_natbib}

\clearpage

\section{Appendix}
\label{sec:appendix}
\subsection{Human-In-The-Loop Details}
\label{sec:appendix-hiip}
\begin{figure*}[htbp]
    \centering
    \includegraphics[width=0.9\textwidth]{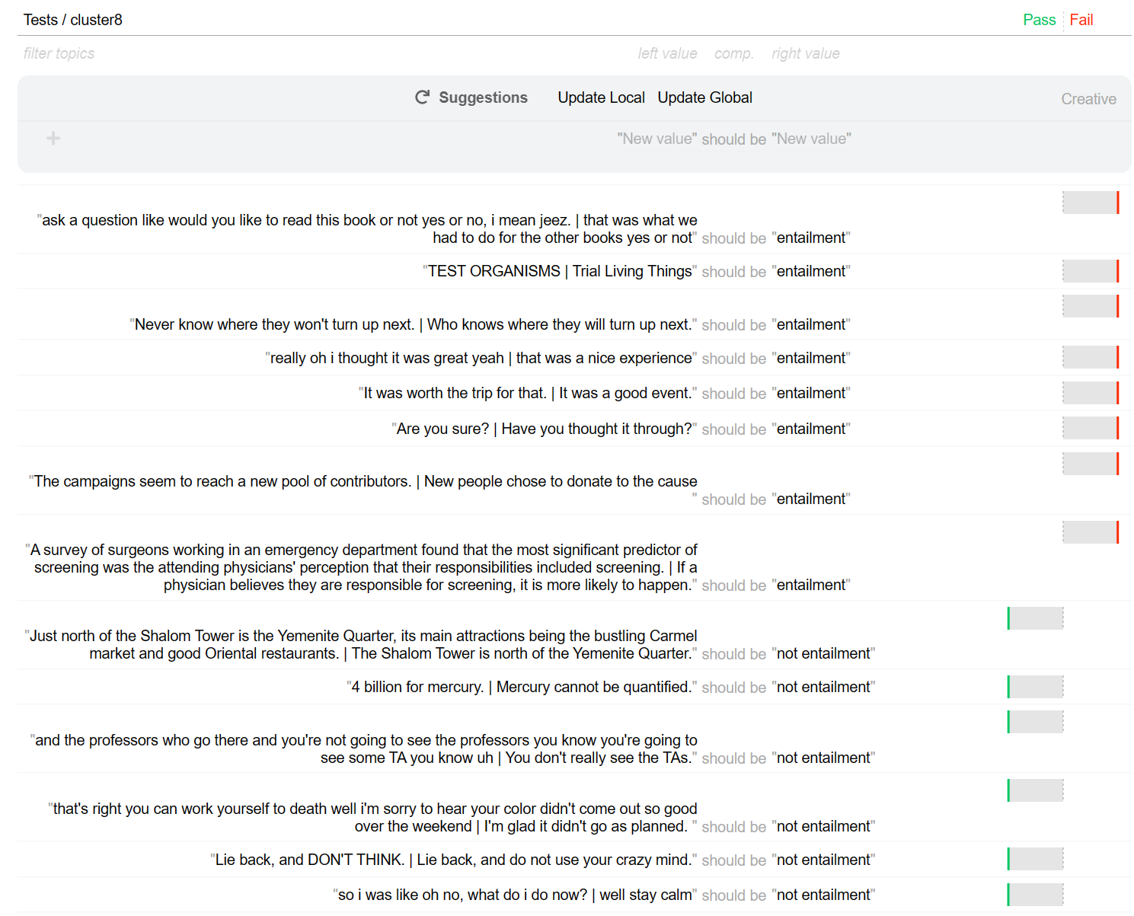}
    \caption{User interface used in our user study}
    \label{fig:ui}
\end{figure*}

\paragraph{User Interface}
The goal of our user study is to find bugs in the target model. To find bugs easier, we provide the following user interface to our users, as shown  in \cref{fig:ui}. The interface is linked with the back-end global and local models.

\begin{figure*}[htbp]
    \centering
    \includegraphics[width=0.9\textwidth]{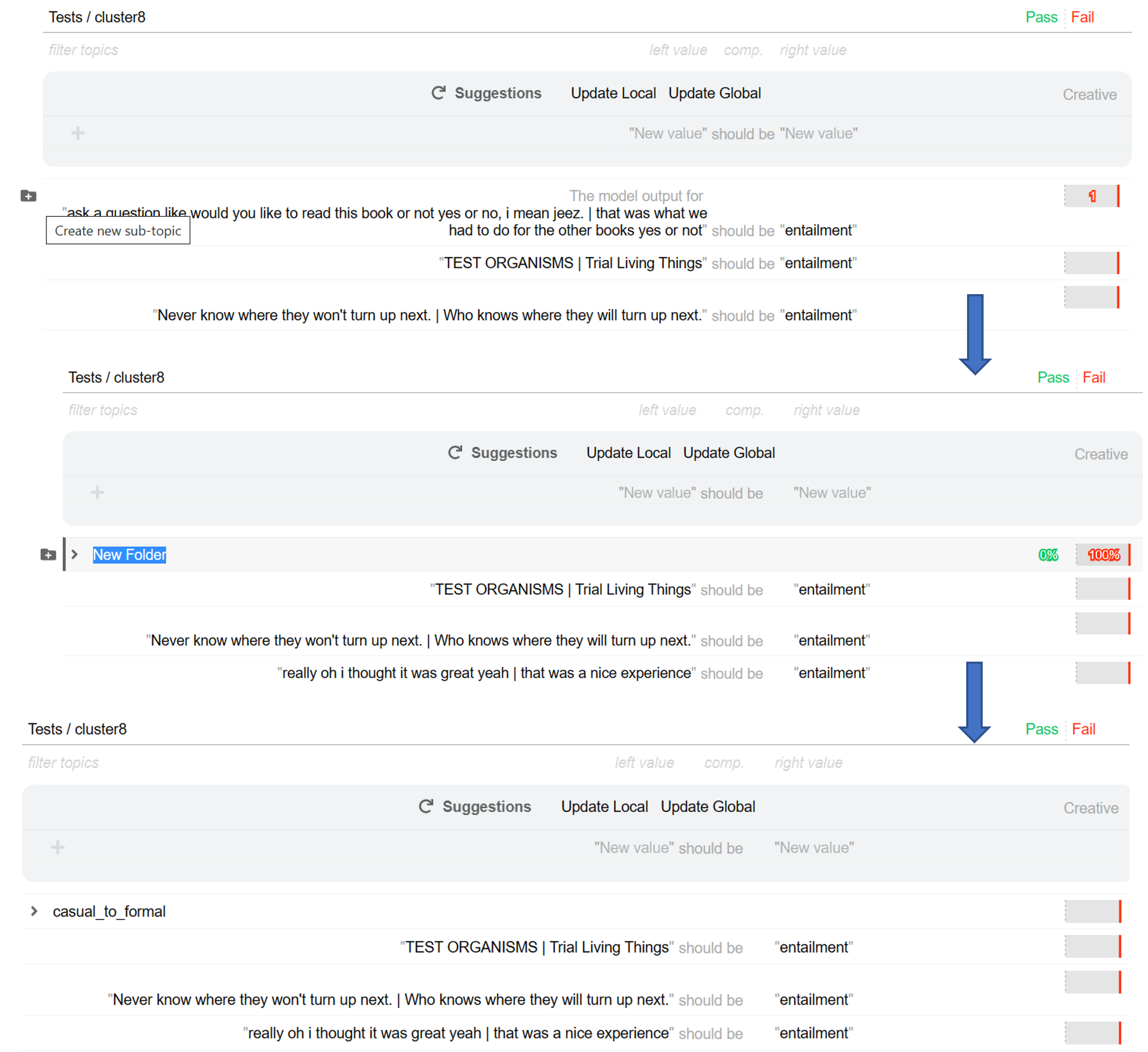}
    \caption{Examples of potential operations.}
    \label{fig:operation}
\end{figure*}

The UI enables the following actions through the bottoms:
\begin{itemize}
    \item Suggest:  click to use the current sentence list as a prompt for GPT-3 to generate similar examples;
    \item Add: allows users to add a sentence from the generated examples to the current list;
    \item Update global: trains the global model using the concatenation of a random sample of sentences from training set and sentences in current list;
    \item Update local: trains the local model using the sentences in the current list,
    \item Creative: indicates whether the local and global models make different decisions. A red color indicates disagreement while green indicates no disagreement.
    \item Rename: Users can rename their clusters to an interpretable name if they'd like to.
\end{itemize}
In \cref{fig:operation}, we show an example of adding a sentence to a subcluster and renaming it.
\paragraph{User Study Introduction}
Our user study consists of two parts. In the first part, users will read the initial sentences displayed on the user interface, which are the clustering results from the TDG automatic subgroup discovery stage. They can further categorize them into smaller sub-clusters if they notice finer-grained groups within the current cluster.

In the second part, users can add more bugs to the cluster or sub-cluster by first clicking on the ``Suggest'' button to request GPT-3 to generate more similar examples. They will then review the suggestions and add valid examples according to the following criteria: (1) if the local model's prediction is incorrect (i.e. the text after ``should be'' is wrong), correct it and add it; or (2) if the global model's prediction differs from the correct local model prediction (i.e. the bar under the ``Creative'' turns red), add it.

We ask that each user clicks on the ``Update global'' button at least once during their study session to ensure that they continue to find meaningful bugs in the updated model.

\subsection{Analysis on Relationships Between Clusters}
\label{sec:cluster-relationship}
We observe that sometimes fine-tuning the model with TDG(all) augmented data on individual clusters can lead to improved performance on certain clusters and worse performance on others. This suggests that there may be relationships between clusters, such as mutual benefit or conflict.

One conjecture is  data points may have multiple patterns shared with different sentences, therefore, belonging to multiple clusters. Each individual TDG is just working on one of them. Combining and fine-tuning together can cumulative the performance. For example,  MNLI example ``\textit{S1: Pray be seated, mademoiselle. S2: Please, everyone be seated.}'' can have both the patterns of the cross-lingual entailment and the monotonicity.
Another conjecture for conflicting clusters is that the patterns within one cluster may be contradictory to those in another cluster. For example, in sentiment classification, sentences mentioning ``American'' in technology topics may conflict with sentences mentioning ``American'' in international relationship topics. 
Such conflicts  may be solved by simply adding  similar examples.
Therefore, fine-tuning these conflicting clusters together may negatively impact the performance of one or both clusters. 
\end{document}